\documentclass[10pt,twocolumn,letterpaper]{article}

\usepackage[pagenumbers]{cvpr} 

\usepackage[dvipsnames]{xcolor}

\usepackage{algorithm}
\usepackage{amsmath}

\newcommand{\R}{\mathbb{R}}

\usepackage{makecell}

\definecolor{cvprblue}{rgb}{0.21,0.49,0.74}
\usepackage[pagebackref,breaklinks,colorlinks,citecolor=cvprblue]{hyperref}

\def\paperID{9313} 
\def\confName{CVPR}
\def\confYear{2024}

\title{Understanding Masked Autoencoders From a Local Contrastive Perspective}

\author{
	Xiaoyu Yue$^{1,2}$
	\;\;Lei Bai\thanks{Corresponding author: bailei@pjlab.org.cn.} $^{ ,2}$
	\;\; Meng Wei$^{2,3}$
	\;\; Jiangmiao Pang$^{2}$
	\;\; Xihui Liu$^{3}$ \\
	Luping Zhou$^{1}$
        \;\; Wanli Ouyang$^{2}$\\
	$^1$The University of Sydney
        \;\; $^2$Shanghai AI Laboratory
        \;\; $^3$The University of Hong Kong
}

\begin{document}
\maketitle
\begin{abstract}
Masked AutoEncoder (MAE) has revolutionized the field of self-supervised learning with its simple yet effective masking and reconstruction strategies.
However, despite achieving state-of-the-art performance across various downstream vision tasks, the underlying mechanisms that drive MAE's efficacy are less well-explored compared to the canonical contrastive learning paradigm.
In this paper, we first propose a local perspective to explicitly extract a local contrastive form from MAE's reconstructive objective at the patch level.
And then we introduce a new empirical framework, called Local Contrastive MAE (LC-MAE), to analyze both reconstructive and contrastive aspects of MAE.
LC-MAE reveals that MAE learns invariance to random masking and ensures distribution consistency between the learned token embeddings and the original images.
Furthermore, we dissect the contribution of the decoder and random masking to MAE's success, revealing both the decoder's learning mechanism and the dual role of random masking as data augmentation and effective receptive field restriction.
Our experimental analysis sheds light on the intricacies of MAE and summarizes some useful design methodologies, which can inspire more powerful visual self-supervised methods.
\end{abstract}
\section{Introduction}
\label{sec:intro}

Recently, self-supervised learning has seen significant progress in the field of computer vision with two dominant paradigms, \textit{i.e.}, Contrastive Learning, and Masked Image Modeling.
The Contrastive Learning methods~\cite{chen2021exploring,he2020momentum,caron2020unsupervised,dwibedi2021little,chen2020simple,grill2020bootstrap,chen2021mocov3,caron2021emerging} benefit from learning invariance by contrasting positive and negative image pairs, which are constructed from random data augmentations.
On the other hand, the Masked Image Modeling paradigm~\cite{bao2021beit,xie2022simmim,he2022masked,gao2022convmae}, which is inspired by Masked Language Modeling in the field of Natural Language Processing, involves randomly masking a portion of an input image and learning to reconstruct the missing pixels based on the visible part.
Recent studies have shown that the ViT features pretrained with Masked Image Modeling have achieved competitive or even better performance than those with Contrastive Learning when finetuning on downstream tasks.  

As a typical MIM method, Masked AutoEncoder (MAE)~\citep{he2022masked} represents a significant breakthrough in visual representation learning, as it paves the way for harnessing the potential of masked autoencoding techniques in vision.
MAE adopts a simple asymmetric encoder-decoder architecture and is pretrained to reconstruct masked images generated through aggressive masking, with a large mask ratio of up to $75\%$.
However, despite the efficiency and simplicity of MAE, the implicit generative way of pretraining presents a challenge to decipher the exact factors contributing to its performance. 
In contrast, the Contrastive Learning paradigm, which has well-defined
formulations and explicit supervision of the encoded features, is more comprehensively understood in the field. 
Given the existence of these two distinct paths in SSL, we believe that gaining a better understanding of MAE can drive SSL toward a unified and more effective direction.

Recent works~\citep{kong2023understanding1, zhang2022mask} have provided a theoretical framework to approximate MAE with contrastive learning, which is promising as it allows for leveraging the insights gained from contrastive learning to interpret MAE. 
While the mathematical approximations and assumptions made in these works provide a good starting point, explicit empirical evaluation at the operational level is still lacking.
To bridge the gap, we first take a novel local perspective to interpret MAE.
Specifically, we propose reformulating MAE's training objective based on image patches, rather than the entire image.
This reformulation enables us to directly decompose MAE's reconstructive objective into a combination of reconstructive and contrastive objectives.
More importantly, based on this decomposition, we propose an empirical framework, namely Local Contrastive MAE (LC-MAE), to analyze both the reconstructive and contrastive aspects of MAE.
LC-MAE not only preserves MAE's state-of-the-art performance on downstream tasks
but also unveils the explicit local contrastive form of MAE.
It uses three explicit token-based loss functions: a reconstruction loss, a cross-view contrastive loss, and an in-view contrastive loss.
The cross-view loss ensures that, for a given image, token features in the same position but different random masks are locally similar, revealing that MAE actually learns invariance to random masking.
The in-view loss ensures the consistency of the distribution of the output features and the input image, which effectively prevents the collapse of MAE.

Apart from connecting MAE to contrastive learning, we also deeply investigate how each component of MAE, \ie, the autoencoder and the random masking, contributes to its success.
For the autoencoder, we focus on analyzing the decoder’s behaviors, and find that the decoder primarily utilizes positional information in the shallow layers, and gradually learns semantic information as going to deeper layers.
This demonstrates that a deep decoder is essential for learning rich semantic representations.
For the random masking strategy, it serves two purposes: 
(1) data augmentation;
(2) restricting the effective receptive field of Vision Transformers, which turns out to be crucial for MAE's effectiveness on downstream tasks.
Surprisingly, we found that solely restricting the receptive field is enough to improve downstream finetuning performance.

We hope our analysis of MAE will inspire future research in the field of visual representation learning.

\section{Related Work}

\textbf{Contrastive learning.}
As the dominant self-supervised representation learning paradigm in the field of computer vision, contrastive learning~\citep{chen2021exploring,he2020momentum,caron2020unsupervised,dwibedi2021little,grill2020bootstrap} learns invariance by comparing random views.
A representative work in this domain is SimCLR~\citep{chen2020simple}, which learns semantic representations by maximizing the similarity between different views derived from the same image within the latent space.
MoCo v3~\citep{chen2021mocov3} explores the pretraining of vision transformers through the methodology of contrastive learning.
DINO~\citep{caron2021emerging} explores new properties of self-supervised vision transformers.

\noindent\textbf{Masked Image Modeling.}
In recent years, the development of Vision Transformers \citep{dosovitskiy2020image,el2021xcit,touvron2021training} has significantly encouraged the application of Masked Image Modeling (MIM).
Originating from Masked Language Modeling, MIM has achieved impressive results in visual self-supervised representation learning.
BEiT~\citep{bao2021beit} maps image patches into visual tokens using d-VAE~\citep{ramesh2021zero} and predicts these visual tokens based on the masked images.
SimMIM~\citep{xie2022simmim} attempts to simplify the algorithmic process of MIM by directly using the original image pixels as the target.
MAE~\citep{he2022masked} employs an encoder-decoder framework to perform image reconstruction tasks.
IBOT~\citep{zhou2021ibot}, CAE~\citep{chen2023context}, and CMAE~\citep{huang2022contrastive} try to combine contrastive learning and MIM.

\noindent\textbf{Understanding MAE.}
Despite the simplicity and efficacy of MAE, there is a paucity of work dedicated to understanding and analyzing its inner mechanism.
Many existing works~\citep{liu2023good,li2022semmae,liu2022mixmim} focus on improving MAE based on intuitive understanding.
\citet{cao2022understand} primarily focuses on the role of self-attention within the MAE framework.
\citet{kong2023understanding} abstracted MAE as a hierarchical latent variable model, thereby analyzing the mechanism through which MAE learns semantic information.
\citet{park2023self}  conducted a comparative analysis of the behavioral differences between the MIM and contrastive learning.
\citet{kong2023understanding1} and \citet{zhang2022mask} reformulate MAE as contrastive learning, sharing similar motivation with us.
However, they both consider masked patches and visible patches as two views for global contrastive learning, while we demonstrate that MAE actually conducts contrastive learning between local regions on the masked image.

\section{A Local Perspective for MAE}

In this section, we introduce the local contrastive learning form of MAE, which is a more accessible way for understanding the intrinsic mechanisms of MAE. 
This contrastive form conceptually elaborates that MAE acquires semantic representations by aligning features at the patch level.
Specifically, we first provide a brief revisit of MAE in Section~\ref{sec:mae_revisit}, and then we show how to reformulate MAE in Section~\ref{sec:mae_reform}.
Finally, in Section~\ref{sec:mae_eval}, we introduce a contrastive learning approach derived from MAE, termed Local Contrastive MAE (LC-MAE).
Empirically, we demonstrate that using only patch-level contrastive losses can also learn features similar to those acquired by masked image modeling.

\subsection{A brief revisit of MAE}
\label{sec:mae_revisit}

\begin{figure*}[t]
    \centering
    \includegraphics[width=0.9\linewidth]{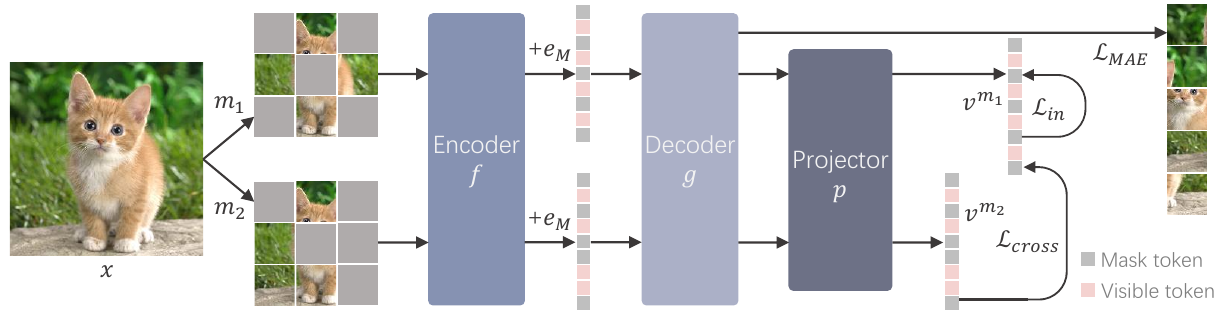}
    \caption{The overall pipeline of LC-MAE. The input image $x$ is masked by $m_1$ and $m_2$, yielding two augmented views. Then the two views are separately fed into the MAE model and projector $p$, resulting in feature vectors $v^{m_1}$ and $v^{m_2}$. In addition to the MAE's reconstruction loss $\mathcal{L}_{MAE}$, two contrastive losses, $\mathcal{L}_{cross}$ and $\mathcal{L}_{in}$, are computed using $v^{m_1}$ and $v^{m_2}$. }
    \label{fig:cl_mae}
\end{figure*}

Masked AutoEncoders (MAE)~\citep{he2022masked} is a straightforward yet efficacious self-supervised method for pretraining Vision Transformers (ViT)~\citep{dosovitskiy2020image,touvron2021training}.
It learns rich hidden representations by masking a large portion of the image and reconstructing the masked patches based on the visible patches.

Formally, given an input image, MAE first partitions it into $n$ non-overlapping patches, denoted as $x = \{ x_i \}_{i=1}^{n}$.
Then, the $n$ patches are split into two complementary subsets with a random binary mask $m \in \{ 0, 1 \}^n$: the \textit{visible patches} $x^v = x[1-m]$ and the \textit{masked patches} $x^m = x[m]$.
MAE adopts an encoder-decoder architecture: $h = g \circ f$.
The encoder $f(\cdot)$ encodes the visible patches into the visible token representations $z$: $z = f(x^v)$.
Then the visible tokens $z$ are fed into the decoder $g(\cdot)$.
Meanwhile, all the masked positions will be filled with a shared learnable mask token $e_M$.
Both $z$ and $e_M$ are added with positional embeddings.
Subsequently, the resulting token sequence is used to reconstruct the original pixels of the mask tokens.
MAE employs the Mean Squared Error (MSE) loss function for pretraining:
\begin{equation}
    \label{eq:mae_loss}
    \begin{aligned}
    \mathcal{L}(x, m) &= \sum_{i=1}^{n} m_i \cdot || h(x, m)_i - x_i ||^2,\\ 
    h(x, m) &= g(f(x[1-m])).
    \end{aligned}
\end{equation}
In the original implementation of MAE, the decoder output is $l_2$-normalized, and a high mask ratio (\eg{} $75\%$) is commonly applied to generate the random mask $m$.

\subsection{Reformulating MAE as Local Contrastive Learning}
\label{sec:mae_reform}
Previous research~\cite{zhang2022mask,kong2023understanding1} has demonstrated that MAE can be better understood from the perspective of contrastive learning.
Although they approximate the formulation of MAE with the global alignment loss in contrastive learning, their interpretation of MAE still relies on certain assumptions.
We find that the MAE naturally maps different masked tokens to similar content, hence a local perspective may be more suitable to describe MAE.

During the entire training process of MAE, for a position $j$ in an input image $x$,   
there exist two different training iterations that construct the masked patch $x_j$ with two randomly generated masks $m_a$ and $m_b$ (where $m_a \neq m_b$).

Disregarding that $m_a$ and $m_b$ are generated in different iterations, we view $m_a$ and $m_b$ as two data augmentations performed on the same image as in contrastive learning.
Hence, from the local perspective (\ie, the image patch level), we rethink the training objective of MAE $\mathcal{L}(x, m)$ \textit{w.r.t.} two random data augmentations $m_a$ and $m_b$ and the $j$-th patch as:
\begin{equation}
\small
\label{eq:two_mask_contrast}
\begin{aligned}
\mathcal{L}(x, m)_j = & || h(x, m_a)_j -x_j ||^2 \\
                   =& ||(h(x, m_a)_j - h(x, m_b)_j) + (h(x, m_b)_j - x_j)||^2 \\
                   =& ||(h(x, m_a)_j - h(x, m_b)_j) + e_{pred}(j | m)||^2,
\end{aligned}
\end{equation}
where $e_{pred}(j | m)$  is the prediction error:
\begin{equation}
\label{eq:recon}
    e_{pred}(j | m) = h(x, m)_j - x_j.
\end{equation}
It represents the reconstruction objective of MAE: the decoder output at the position $j$ with random masking operation $m$ should always approximate the invariant image patch value $x_j$.

The first term of Eq.~\ref{eq:two_mask_contrast} $(h(x, m_a)_j - h(x, m_b)_j)$ can be regarded as a \textit{patch-level contrastive objective}: align the local patch features in two augmented views.

Considering that Eq.~\ref{eq:two_mask_contrast} is just a conceptual reformulation of MAE, as the two random augmentations $m_a$ and $m_b$ do not actually occur in the same training iteration, we further provide a reformulation based on the masked image patches in a single forward of MAE.

Similarly, we rethink $\mathcal{L}(x, m)$ \textit{w.r.t.} two different masked positions $i$ and $j$ of an input image $x$ as:
\begin{equation}
\label{eq:one_mask_contrast}
\begin{aligned}
    \mathcal{L}(x, m)_i =& || h(x, m)_i - x_j + x_j - x_i ||^2 \\ 
\end{aligned}
\end{equation}
Substituting $x_j$ with $h(x, m)_j - e_{pred}(j | m)$ from Eq.~\ref{eq:recon}
\[ \begin{aligned}
    \mathcal{L}(x, m)_i =& || (h(x, m)_i - (h(x, m)_j - e_{pred}(j|m)) \\
        &- (x_i - x_j))||^2\\
        =& || \left [ (h(x, m)_i - h(x, m)_j) - (x_i - x_j) \right ] \\
        &+ e_{pred}(j|m)||^2 .
\end{aligned} \]
The first term requires that the difference between the decoder outputs at positions $i$ and $j$, \ie, $(h(x, m)_i - h(x, m)_j)$, should approximate their pixel difference $(x_i - x_j)$.
This can also be regarded as the contrastive objective at the patch level with the distance between the masked patches as an additional margin.

So far, we have separated the local contrastive form of MAE from its reconstructive form, in both Eq.~\ref{eq:two_mask_contrast} and Eq.~\ref{eq:one_mask_contrast}.
Just as pointed out in \citet{zhang2022mask}, MAE's reconstruction ability is not enough to explain the efficacy of MAE, since pretraining with vanilla autoencoder performs much worse than MAE.
Hence, we believe that the derived local contrastive form, which indicates that \textit{MAE implicitly aligns local features}, may help uncover its underlying mechanism.

In the following section, we will introduce an architecture that transforms the implicit local contrastive form of MAE into an explicit form, in order to empirically evaluate the impact of each aspect of MAE.

\subsection{Explicit Local Contrastive Form of MAE}

\label{sec:mae_eval}
Based on Eq.~\ref{eq:two_mask_contrast} and Eq.~\ref{eq:one_mask_contrast}, we summarize three possible training objectives of MAE:
\begin{enumerate}[label=\roman*]
    \item Patch reconstruction.
    \item The first term of Eq.~\ref{eq:two_mask_contrast}, aims to minimize the patch token embedding distance in different views of an image.
    \item The first term of Eq.~\ref{eq:one_mask_contrast}, aims to minimize the discrepancy between the distribution of the output token embeddings and the distribution of the input image patches.
\end{enumerate}
However, the framework of MAE makes it impossible to explicitly perform ablative studies on each individual training objective.
Hence, we design a simple framework to instantiate the local contrastive learning form of MAE into explicit loss functions, namely Local Contrastive MAE (LC-MAE).

The overall pipeline of LC-MAE is shown in Figure~\ref{fig:cl_mae}. 
Compared to MAE, the input of LC-MAE becomes two masked views of the same image.
It has three loss functions: 
the reconstruction loss $\mathcal{L}_{MAE}$ (\textit{w.r.t.} objective i);
the cross-view loss $\mathcal{L}_{cross}$ (\textit{w.r.t.} objective ii);
and the in-view loss $\mathcal{L}_{in}$ (\textit{w.r.t.} objective iii).
LC-MAE employs the same encoder and decoder architecture as MAE.
Hence, the reconstruction loss $\mathcal{L}_{MAE}$ remains consistent with that of MAE.
For calculating contrastive losses $\mathcal{L}_{in}$ and $\mathcal{L}_{cross}$., we introduce an extra fully connected layer as the projector ($p(\cdot)$) to produce the output vector $v = p(h(x, m))$.


In the application of the cross-view loss $\mathcal{L}_{cross}$, the model utilizes two random masks, $m_1$ and $m_2$, to mask the input image $x$. The resulting output vectors for the two masked views are denoted as $v^{m_1}$ and $v^{m_2}$, respectively. 
$\mathcal{L}_{cross}$ maximizes the cosine similarity between the paired vectors at positions that are simultaneously masked in two views.
This operation is equivalent to the mean squared error of $l_2$-normalized vectors~\cite{grill2020bootstrap}, formally:
\begin{equation}
    \mathcal{L}_{cross} = \sum_i^n (m_{1,i} \cdot  m_{2,i}) \cdot ( 1 - cos(v^{m_1}_i, v^{m_2}_i)),
\end{equation}
where $cos(\cdot, \cdot)$ denotes the cosine similarity, and $i$ is the patch index.

The in-view loss $\mathcal{L}_{in}$ is calculated within a single masked view.
It leverages the cosine similarity matrix of the input image patches $x$ as the supervision for the output tokens, in order to explicitly align these two distributions.
The mean absolute error between the cosine similarity matrix of output tokens $v$ and the original image patches is calculated as:
\begin{equation}
    \mathcal{L}_{in} = \sum_{i \neq j}^n (m_i \cdot m_j) \cdot | cos(v_i, v_j) - cos(x_i, x_j) |.
\end{equation}
The overall loss of LC-MAE is the sum of these three losses: $\mathcal{L}_{MAE} + \mathcal{L}_{in} + \mathcal{L}_{cross}$.

We pretrain the LC-MAE on ImageNet with a mask ratio of $75\%$ for 100 epochs.
And then supervised finetuning the model on ImageNet for an additional 100 epochs.
To shorten the pretraining duration, we set the depth of the decoder to 2.

\noindent\textbf{Ablation Study.} We conducted ablation studies to demonstrate the effectiveness of each loss function and analyzed the key success
factors of MAE.
As shown in Table~\ref{tab:lc_mae}: 

\noindent(1) Setting (a) is LC-MAE, which incorporates all three losses, showing a slight improvement compared to Setting (b), \ie, the original MAE (reconstruction loss only). 
This indicates that LC-MAE retains the effectiveness of MAE.
\begin{table}
  \centering
  \begin{tabular}{c l l l c}
    \toprule
    Setting & $\mathcal{L}_{MAE}$ & $\mathcal{L}_{cross}$ & $\mathcal{L}_{in}$ & FT Acc(\%) \\
    \midrule
    (a) & \checkmark & \checkmark & \checkmark & 83.0 \\
    (b) & \checkmark & & & 82.9 \\
    (c) & & \checkmark & & Collapse \\
    (d) & & & \checkmark & 82.2 \\
    (e) & \checkmark & \checkmark & & 83.1 \\
    (f) & \checkmark & & \checkmark & 82.8 \\
    (g) &  & \checkmark & \checkmark & 82.5 \\
    \bottomrule
  \end{tabular}
  \caption{Finetuning accuracy (FT Acc) of LC-MAE under different settings.}
  \label{tab:lc_mae}
\end{table}


\noindent(2) The reconstruction loss consistently improves performance, with gains of $0.5\%$ in setting (a) versus (g), and $0.6\%$ in setting (f) versus (d).
Hence, the reconstruction loss is indeed simple yet effective.


\noindent(3) The cross-view loss leads to performance gains of $0.3\%$ in setting (g) versus (d), and $0.2\%$ in setting (a) versus (f).
In MAE, the cross-view loss is implicitly realized during the training process, while in LC-MAE, its explicit implementation enhances the capacity to learn semantic representations.

\noindent(4) The in-view loss leads to a slight performance decline of $0.1\%$, possibly due to a suboptimal similarity matrix.

\noindent(5) It is noteworthy that setting (g), even without the use of reconstruction loss, still achieves a finetuning accuracy of $82.5\%$, corroborating the intrinsic nature of contrastive learning implicitly performed by MAE, and hinting at the potential of pure contrastive learning.

\noindent\textbf{How MAE avoids Feature Collapse.} 
From the ablation study, we can also explain why MAE doesn't suffer from feature collapse when only aligning the positive pairs. 
Since LC-MAE inherits the network architecture of MAE and does not employ techniques such as extra predictor, momentum encoder, or stop-gradient to prevent model collapse, using only the cross-view loss (setting (c) in Table~\ref{tab:lc_mae}) leads to collapse into a constant point.
When the model is equipped with the in-view loss (setting (g)) or reconstruction loss (setting (e)), it is able to avoid collapsed solutions.
This indicates that MAE actually employs two methods to avoid collapse: 1) assigning a unique target value for each predicted value; and 2) ensuring that the output has a distribution similar to that of the input image.
In both methods, collapse would result in large loss values.



\section{Component Analysis of MAE}

In this section, based on the reformulation of MAE, we dissect the role of two main components of MAE: autoencoder and masking, and delve into what contributes to its success as a self-supervised framework.
Within the autoencoder, the encoder serves as the network utilized for downstream tasks, whereas the decoder, which reconstructs masked tokens back into pixels, is the most innovative part of MAE.
For this reason, we focus on the significance of the decoder, and explore the decoding process in Section~\ref{sec:dec}, unveiling how MAE learns semantic features and conducts local contrastive learning.
Subsequently, the role of masking is discussed in Section~\ref{sec:mask}, where we discover that aside from serving as data augmentation and providing a learning target for MAE, masking also effectively controls the network's effective receptive field, which is one of the reasons for MAE's strong performance in downstream tasks.

\subsection{Decoder}
\label{sec:dec}

To uncover the inner mechanisms of MAE, it's critical to comprehend the decoder's role in helping the encoder learn rich hidden representations in a generative manner, even though the decoder will be discarded after pertaining.

\begin{figure}
    \centering
    \includegraphics[width=\linewidth]{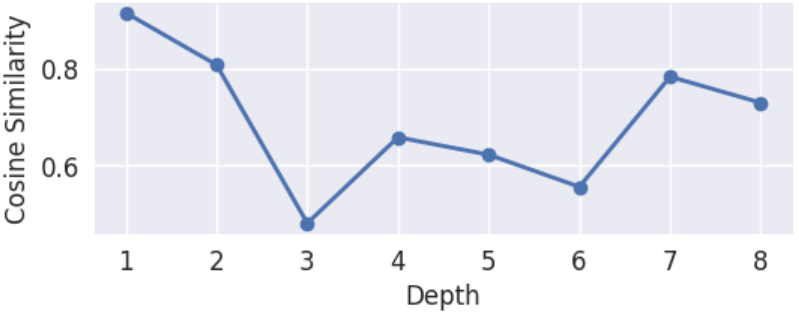}
    \caption{The average attention similarity for each decoder layer, indicates the degree of reliance on positional information.}
    \label{fig:dec_attn_cos_sim}
\end{figure}

We noticed that the mask token $e_M$ is shared across all masked positions, with only the added position embeddings varying.
Nevertheless, considering the distinctive content decoded from various masked positions, we speculate that the decoding process of MAE might be initially guided by positional information.

To examine this assumption, we conduct statistical analysis on the attention maps from different decoder layers using the validation set of ImageNet-1K.
To reduce the complexity of the analysis, we deliberately mask all the images with identical random binary masks (\ie, masked positions are kept the same).
Let $\mathcal{I}$ denote the set of all images, for the $l$-$th$ decoder layer, we extract the attention maps from all masked positions of the $i$-$th$ image, defined as $A_{l, i} \in \R^{k \times h \times n}$ with $k$, $h$, and $n$ denote the number of mask tokens, the number of heads, and the number of patches, respectively.
Then we compute the cosine similarity of each attention map pair $cos(A_{l, i}, A_{l, j})$ and average the similarity scores across the whole image set:
\begin{equation}
    S^{attn}_l = \frac{\sum_{i \neq j}^{|\mathcal{I}|}cos(A_{l, i}, A_{l, j})}{|\mathcal{I}|(|\mathcal{I}| - 1)},
\end{equation}
where $|\mathcal{I}|$ is the number of images, and $S^{attn}_l$ is the average cosine similarity of attention maps at $l$-$th$ decoder layer.
A higher similarity means the decoder layer relies more on invariant features shared across images, \ie, the positional information. 

As shown in Figure~\ref{fig:dec_attn_cos_sim},  the average attention map similarity across all images is the highest at the first decoder layer (up to $0.9$) and lowest at the third layer.
The average cosine similarity of the first two layers is significantly higher than that of the subsequent layers.
This suggests that \textbf{ positional information is learned primarily at the shallower layers.}

\begin{figure}
    \centering
    \includegraphics[width=\linewidth]{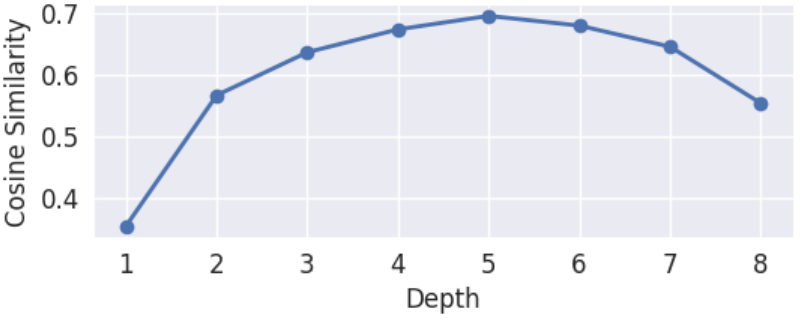}
    \caption{The average cosine similarity between the DINO features and the features of each decoder layer, quantifies the semantic richness of decoder features.}
    \label{fig:dec_attn_dino_sim}
\end{figure}

\begin{figure*}
    \centering
    \includegraphics[width=\linewidth]{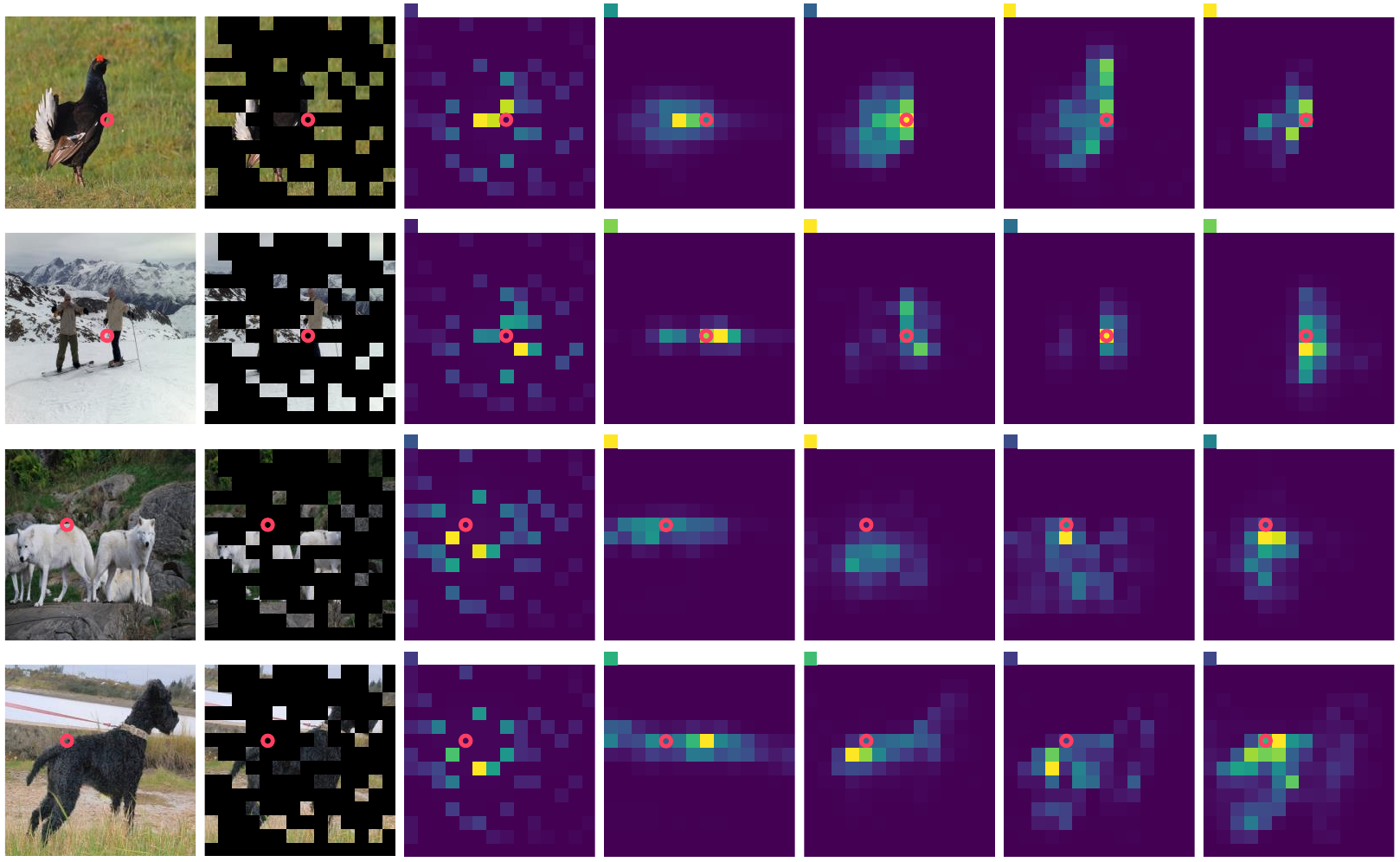}
    \caption{Visualization of attention maps from decoder layers.
    From left to right: input images, masked images, and attention maps of the 1-st, 2-nd, 3-rd, 5-th, 8-th decoder layers. The red circles (\textcolor{red}{$\bigcirc$}) denote the masked tokens serving as queries.
    The first two rows and the last two rows show mask tokens at different positions.
    The block in the top-left of each attention map is the weight for the [CLS] token.}
    \label{fig:dec_attn}
\end{figure*}

Then the intriguing question is: \textit{How does the decoder acquire the semantic information?}
Inspired by CutLER~\cite{wang2023cut}, LOST~\cite{simeoni2021localizing}, and TokenCut~\cite{wang2023tokencut}, we use features extracted by DINO~\cite{caron2021emerging} as an indicator to measure the semantic richness of each decoder layer.
DINO~\cite{caron2021emerging} is a widely adopted self-supervised vision model, which produces features that appear to contain explicit information about semantic segmentation.
Following CutLER~\cite{wang2023cut}, we calculate the cosine similarity between patch features within the same layer and use the normalized cosine similarity matrix as the similarity weight $W$, which is a $n \times n$ symmetrical matrix.
Then, by computing the average cosine similarity between the similarity weights of each decoder layer and those from the final layer of DINO, we can quantify the extent to which the decoder utilizes semantic information. Formally:
\begin{equation}
    S^{sim}_l = \frac{\sum_{i}^{|\mathcal{I}|}cos(W^{dino}_{i}, W^{dec}_{l,i})}{|\mathcal{I}|},
\end{equation}
where $i$ is the index of images, $W^{dino}$ and $W^{dec}$ denote the similarity weights of features from DINO and decoder layers respectively, and $S^{sim}_l$ is the average cosine similarity between them at the $l$-$th$ decoder layer.
$S^{sim}$ is depicted in Figure~\ref{fig:dec_attn_dino_sim}, one can find that the similarity is lower for the shallow layers and higher for deep layers, indicating that \textbf{deeper layers are more conducive to learning semantic information.}
\\
\\
\noindent\textbf{Qualitative Analysis}. 
To provide further insights, we visualize the attention maps from decoder layers in Figure~\ref{fig:dec_attn}.
In the first decoder layer, where $S^{attn}$ is the highest and $S^{sim}$ is the lowest, the attention maps exhibit very similar patterns for the very different images.
Starting from the second layer, a gradual shift in attention behavior can be observed. As $S^{attn}$ begins to decrease and $S^{sim}$ increases, the attention maps demonstrate a transition from focusing on fixed positions to adjacent foreground objects.
This conclusion underscores the significance of a deep decoder for the efficacy of MAE's pretraining, and aligns with the results of the ablation study about the decoder's depth conducted in the original MAE paper~\cite{he2022masked} which demonstrated that a deeper decoder outperforms a shallower one in linear probing.
\begin{table}
  \centering
  \begin{tabular}{l c}
    \toprule
    Decoder  & FT Acc(\%)\\
    \midrule
    Transformer & 82.9 \\
    Weighted Average & 82.5 \\
    Conv Layer & 82.9 \\
    \bottomrule
  \end{tabular}
  \caption{Comparisons of different decoders. As the decoder, single-layer convolution and weighted average exhibit effects akin to the transformer.}
  \label{tab:decoder_type}
\end{table}

\noindent\textit{\textbf{The decoder mainly utilizes local features.}}
From Figure~\ref{fig:dec_attn}, another noteworthy observation is that the attention maps of mask tokens tend to focus more on nearby tokens.
This implies that the decoder relies less on global image information for reconstruction, suggesting that local features are adequate for MAE's efficacy.

\noindent\textbf{Empirical Validation}. To validate this hypothesis, we replace the transformer-based decoder with operations that explicitly have a limited receptive field.
Our first attempt is a nonparametric weighted average operation, in which the weights are set as a normalized two-dimensional Gaussian, with $\sigma = 1$ and the size of the receptive field is about $5 \times 5$.
A multi-layer perception (MLP) block is adopted to predict the masked pixels.
We pretrain this \textit{\textbf{Weighted Average}} decoder version and the original \textit{\textbf{Transformer}} decoder version of MAE on ImageNet-1K~\citep{russakovsky2015imagenet} for 100 epochs using the same training strategy.
As shown in Table~\ref{tab:decoder_type}, surprisingly, the \textit{\textbf{Weighted Average}} decoder achieved a finetuning accuracy of 82.5$\%$, which is only 0.4$\%$ lower than the \textit{\textbf{Transformer}} decoder with much fewer parameters.
Furthermore, we employ a convolutional layer with a kernel size of 5 as the decoder.
We can see that this \textit{\textbf{Conv Layer}} decoder achieves the same finetuning accuracy as the original MAE, reaching 82.9$\%$.
The result aligns with the previous observations and provides support for our interpretation of MAE's decoder behavior: \textit{local features within a limited region is sufficient for the decoder to reconstruct the image.}

\subsection{Random Masking}
\label{sec:mask}

In the local contrastive learning form of MAE, random masking acts as data augmentation, enabling MAE to learn invariance to missing image patches.
From this perspective, the role of random masking is akin to random erasing~\cite{zhong2020random}.
However, even with complex self-supervised learning techniques, random erasing struggles to achieve the same level of effectiveness as random masking.
It suggests that random masking provides additional benefits to MAE beyond just data augmentation.\\

\noindent\textbf{(1) Mask Ratio Controls The Effective Receptive Field.}\\
We get started by considering the decoder's attention maps.
When without random masking, the most useful feature for reconstructing a patch's pixels would be its own token.
In this situation, the decoder would have no need to attend to other tokens.
It is the random masking that forces the decoder to expand its receptive field to gather sufficient information for masked patch prediction.

\begin{figure}
    \centering
    \includegraphics[width=\linewidth]{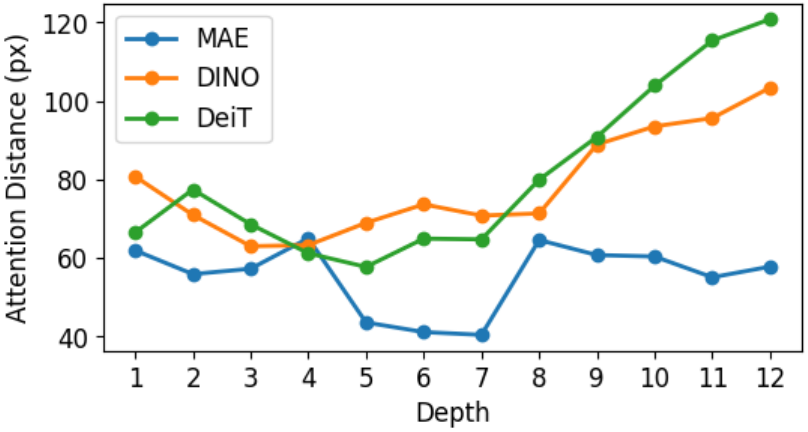}
    \caption{The attention distance for MAE, DINO, and DeiT. MAE's attention distance is significantly smaller than that of the other pretraining methods.}
    \label{fig:attn_dist_pretrain}
\end{figure}

\begin{figure}
    \centering
    \includegraphics[width=\linewidth]{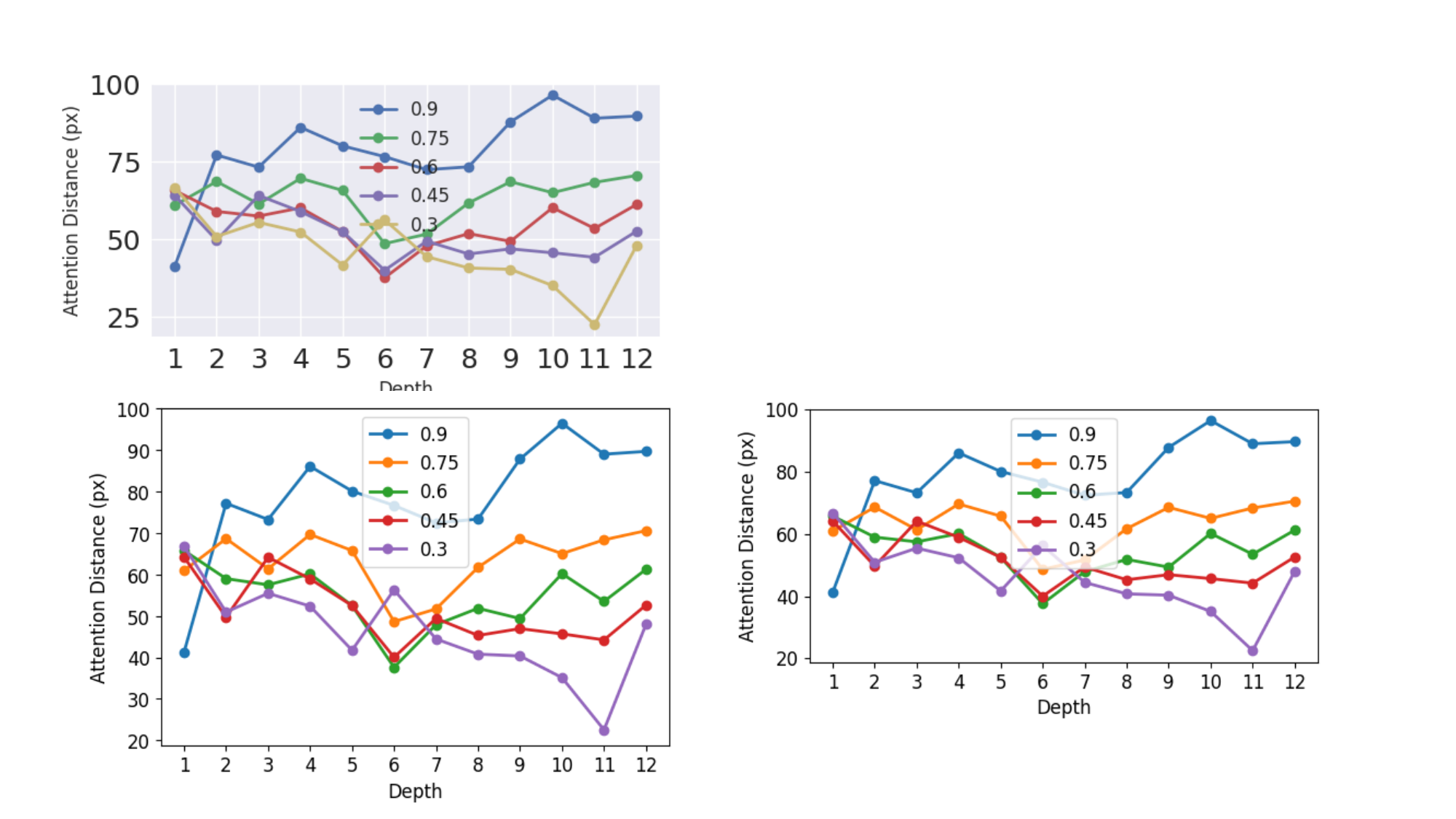}
    \caption{The attention distance for MAE across different mask ratios. There is a significant correlation between mask ratio and attention distance, a larger mask ratio leads to a greater attention distance.}
    \label{fig:attn_dist_mr}
\end{figure}

Furthermore, the restricted receptive field of the decoder requires that the features output by the encoder are robust within local regions, implying that the encoder also primarily focuses on local features.
To verify this, we compare the attention distance of MAE~\cite{he2022masked}, DINO~\cite{caron2021emerging}, and supervised pretrained DeiT\cite{touvron2021training}.
Attention distance \cite{dosovitskiy2020image} is defined as the average distance between the query tokens and key tokens, multiplied by the attention weights.
It is conceptually similar to the size of the effective receptive fields in CNNs.
As illustrated in Figure~\ref{fig:attn_dist_pretrain}, the attention distance of MAE is significantly lower than that of the contrastive learning method DINO and supervised pretrained DeiT, which validates the emphasis of MAEs on local information.

The intensity of random masking is controlled by the mask ratio.
Intuitively, a smaller mask ratio permits the decoder to find features helpful for reconstruction within a narrower range, thereby indirectly controlling the size of the region for contrastive learning.
We pretrain MAE on ImageNet-1K~\citep{russakovsky2015imagenet} for 100 epochs with mask ratios of $0.9$, $0.75$, $0.6$, $0.45$, and $0.3$.
The attention distances for different mask ratios are depicted in Figure~\ref{fig:attn_dist_mr}.
As we hypothesized, the mask ratio and attention distance exhibit a strong correlation, with the impact of mask ratio on attention distance being more pronounced in the deeper layers of the network.
Specifically, a larger mask ratio results in a larger effective receptive field.\\

\noindent\textbf{(2) Limited Receptive Field Benefits Downstream Tasks.}\\
As we discussed above, the role of random masking is twofold: acting as data augmentation and restricting the receptive field.
In this part, we aim to clarify that the limited effective receptive field induced by random masking contributes to the success of MAE on downstream tasks.
In fact, previous works~\cite{yuan2021incorporating,liu2021swin,wu2021cvt} have demonstrated that constraining the receptive field of vision transformers can accelerate convergence and enhance classification performance.

\begin{table}
  \centering
  \begin{tabular}{l c c}
    \toprule
    Pretrain Methods  & FT Acc(\%) & LP Acc (\%)\\
    \midrule
    Random Init & 78.6 & - \\
    MAE & 82.9 & 55.4 \\
    \midrule
    AE(16) & 78.7 & 5.7 \\
    AE(48) & 81.2 & 5.7 \\
    AE(80) & 81.5 & 5.6 \\
    AE(112) & 81.8 & 5.6 \\
    \bottomrule
  \end{tabular}
  \caption{Finetuning accuracy (FT Acc) and linear probing accuracy (LP Acc) of ViT-B/16 pretrained by MAE and AutoEncoder (AE) on ImageNet-1K. The number in parentheses indicates the size of the image block each token reconstructs.}
  \label{tab:autoencoder}
\end{table}

\begin{figure}
    \centering
    \includegraphics[width=\linewidth]{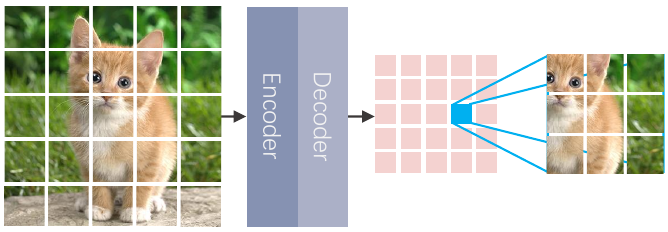}
    \caption{The overall architecture of autoencoder. Each token predicts an image block centered around itself. The image patches on the right are the ground truth for the token marked in blue.}
    \label{fig:ae}
\end{figure}

\noindent\textbf{Empirical Validation}.
To isolate the effect of random masking as data augmentation and solely control the network's receptive field,
we make several modifications to MAE: 1) We discard the random masking operation, which degrades MAE to canonical AutoEncoder (AE); 2) We require each token to predict the pixel values of an image block centered around itself, which requires each token to focus on its surroundings, thus controlling the size of the receptive field.
The overall architecture is shown in Figure~\ref{fig:ae}.

We pretrain MAE and autoencoders on ImageNet-1K and evaluate their finetuning accuracy as well as their linear probing accuracy.
The results are shown in Table~\ref{tab:autoencoder}.
The patch size for ViT models is 16, so AE(16) represents MAE without random masking, while other versions of AE are autoencoders that have an increased size of image blocks to be predicted.
Due to computational resource limitations, we do not use image block sizes larger than 112.
When random masking is simply removed, the finetuning accuracy of MAE drops from $82.9\%$ to $78.7\%$, which is similar to random initialization.
When we enlarge the size of the target image block, the finetuning accuracy dramatically increases to $81.2\%$ and peaks with a block size of 112, reaching $81.8\%$.
This suggests that the excellent finetuning accuracy of MAE is largely attributable to the appropriate receptive field afforded by random masking.
However, the linear probing performance of all the autoencoders is poor, amounting to only about one-tenth of that achieved by the MAE ($5.6\%$ \vs $55.4\%$), indicating that the absence of random masking prevents the network from effectively learning semantic information.

In summary, random masking during pretraining \textit{\textbf{(1) augments training samples, enabling MAE to learn semantically meaningful representations,}} and \textit{\textbf{(2) restricts the size of effective receptive fields, leading to  MAE's strong finetuning performance on downstream tasks.}}

\section{Experimental Details}

\noindent\textbf{For pretraining,} we conduct experiments on the ImageNet-1K~\citep{russakovsky2015imagenet} training set, with ViT-B/16~\citep{touvron2021training} employed as the default backbone.
For MAE pretraining, the mask ratio is set to 0.75 by default.
We use AdamW~\citep{loshchilov2017decoupled} as the optimizer, with a batch size of 1024.
The base learning rate $base\_lr$ is initialized as 1.5e-4, and the actual learning rate $lr = base\_lr \times \frac{\text{batch\_size}}{256}$.
We adopt a 20-epoch linear warmup, and then the learning rate decays with the cosine scheduler~\citep{loshchilov2016sgdr}.
For the pretraining of LC-MAE, we set the base learning rate to 6e-4. The output dimension of the projector is set to 1024 by default.

\noindent\textbf{For finetuning,} all pretrained models employ the same strategy.
We use the AdamW optimizer with a cosine decay learning rate schedule.
The total training epoch number and the batch size are set to 100 and 1024, respectively.
We set the base learning rate as 1.0e-3, and use a 5-epoch warmup.

\section{Conclusion}

In this paper, we propose to understand MAE from a local perspective and introduce the local contrastive learning form of MAE.
We design an empirical framework, Local Contrastive MAE (LC-MAE), to instantiate the contrastive learning form of MAE, which explicitly decomposes MAE's training objective into a reconstruction loss and two contrastive losses in cross-view and in-view.
With the two contrastive losses, we find that MAE inherently learns invariance to random masking and prevents collapse by constraining the distribution of the output feature vectors.
We also delve into the roles of the two main components of MAE—decoder and random masking—in pretraining and summarize several useful designs, including (1) a deep decoder facilitates the learning of semantic information; (2) an appropriate receptive field size can improve finetuning accuracy for downstream tasks.
We hope our findings can inspire future works to design
more powerful self-supervised methods.

{
    \small
    \bibliographystyle{ieeenat_fullname}
    \bibliography{main}
}

\clearpage
\setcounter{page}{1}

\counterwithin{figure}{section}
\counterwithin{table}{section}

\renewcommand\thesection{\Alph{section}}

\setcounter{section}{0}

\setcounter{figure}{0}
\setcounter{table}{0}

\section{Additional Details of Autoencoders}

One of the most important conclusions of our work is that the excellent finetuning performance of MAE is largely attribute to the limited receptive field brought by random masking.
To substantiate this, in Table 3, we remove the random masking operation from MAE and attempt to control the size of the effective receptive field by adjusting the size of the predicted image blocks.
The rationale behind this is that if the autoencoder aims to reconstruct the pixel values of an image block, its effective receptive field should at least cover this area.

\begin{figure}
    \centering
    \includegraphics[width=\linewidth]{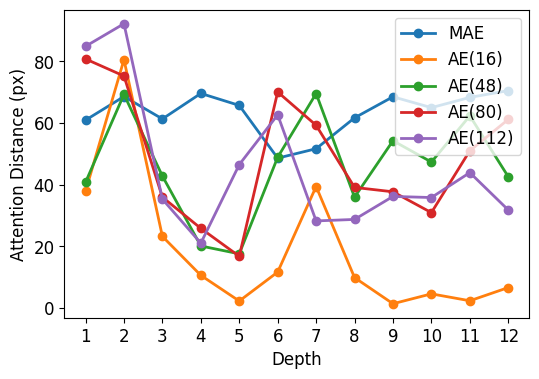}
    \caption{The attention distance for MAE and AEs with different prediction sizes.}
    \label{fig:attn_dist_ae}
\end{figure}

In Figure~\ref{fig:attn_dist_ae}, we present the attention distance for autoencoders with different prediction sizes.
It can be observed that the attention distance of the autoencoder with a prediction size of 16 is significantly smaller than that of other models.
Meanwhile, by controlling the size of image blocks, the effective receptive field of the autoencoder can be effectively expanded.

However, the attention distances of the autoencoder and MAE exhibit different patterns: the attention distance of each layer in MAE is more stable, while autoencoders tend to learn a larger receptive field in the first two layers.
This is because autoencoders have access to all patches, making it more efficient to expand the receptive field at the shallower layers.
To further verify the contribution of restricting the receptive field to finetuning performance, following \citet{asano2019critical} and \citet{kong2023understanding1}, we pretrain MAE using only a single image to limit its capacity to learn semantic information.
The results, as shown in Table~\ref{tab:mae_one_image}, show that even when pretrained using a single image for 5 epochs, MAE achieves a finetuning accuracy of $81.8\%$, which is significantly higher than random initialization.
\citet{kong2023understanding1} suggests that MAE learns a form of data-agnostic favored initialization,  and we believe that this initialization is the restriction of the receptive field.

\begin{table}[b]
  \centering
  \begin{tabular}{l c c}
    \toprule
    Pretrain Images & Pretrain Epochs  & FT Acc(\%)\\
    \midrule
    1 & 5 & 81.8 \\
    ImageNet & 100 & 82.9 \\
    \midrule
    Random Init & - & 78.6 \\
    \bottomrule
  \end{tabular}
  \caption{Comparisons of MAE pretrained with different numbers of images. ``ImageNet'' means using the whole training set of ImageNet for pretraining.}
  \label{tab:mae_one_image}
\end{table}

\section{More Visualizations}

\begin{figure*}
    \centering
    \includegraphics[width=\linewidth]{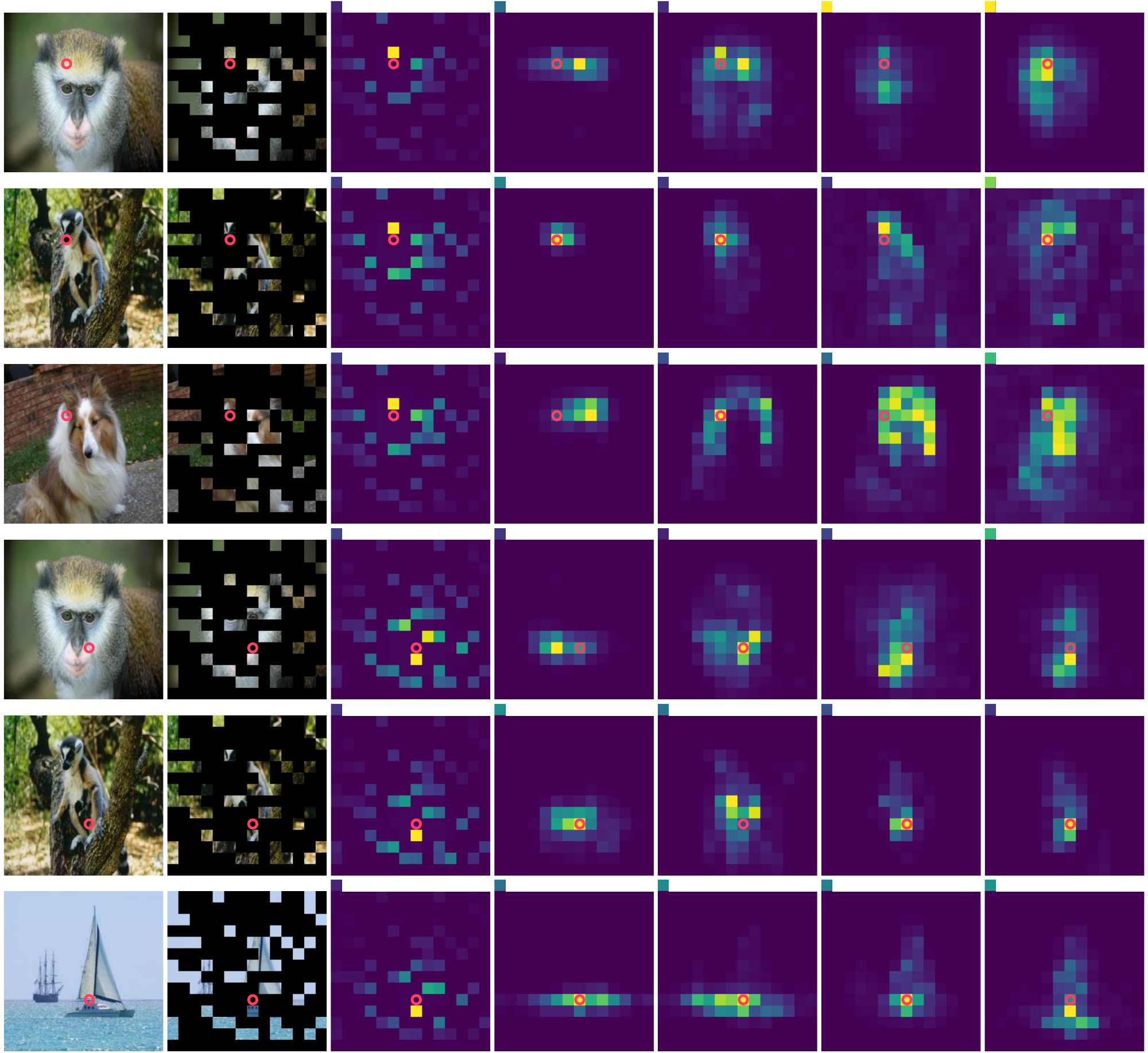}
    \caption{Visualization of attention maps from decoder layers.
    From left to right: input images, masked images, and attention maps of the 1-st, 2-nd, 3-rd, 5-th, 8-th decoder layers. The red circles (\textcolor{red}{$\bigcirc$}) denote the masked tokens serving as queries.
    The block in the top-left of each attention map is the weight for the [CLS] token.}
    \label{fig:more_dec_attn}
\end{figure*}

We provide more visualization of attention maps from decoder layers as shown in Figure~\ref{fig:more_dec_attn}.
As discussed in Section 4.1, the attention maps of the first decoder layer exhibit strong positional relevance, primarily focusing on the visible tokens surrounding the query token.
The second layer shows a transition towards semantic relevance, and subsequent layers mainly focus on semantically related areas.
Comparing different masked positions of the same image (first row \vs fourth row, second row \vs fifth row), we find that masked tokens pay more attention to their own surrounding local regions.

\end{document}